# Robust Pre-Training of Medical Vision-and-Language Models with Domain-Invariant Multi-Modal Masked Reconstruction

Melika Filvantorkaman[1,*], Mohsen Piri[2]

[1]Department of Electrical and Computer Engineering, University of Rochester, Rochester, NY 14627, United States

[2]Department of Electronic, College of Engineering, Kermanshah Science and Research Branch, Islamic Azad University, Kermanshah, Iran

[*]Corresponding author: mfilvant@ur.rochester.edu

## Abstract

Medical vision-language models show strong potential for joint reasoning over medical images and clinical text, but their performance often degrades under domain shift caused by variations in imaging devices, acquisition protocols, and reporting styles. Existing multi-modal pre-training methods largely overlook robustness, treating it as a downstream adaptation problem. In this work, we propose Robust Multi-Modal Masked Reconstruction (Robust-MMR), a self-supervised pre-training framework that explicitly incorporates robustness objectives into masked vision-language learning. Robust-MMR integrates asymmetric perturbation-aware masking, domain-consistency regularization, and modality-resilience constraints to encourage domain-invariant representations. We evaluate Robust-MMR on multiple medical vision-language benchmarks, including medical visual question answering (VQA-RAD, SLAKE, VQA-2019), cross-domain image-text classification (MELINDA), and robust image-caption retrieval (ROCO). Robust-MMR achieves 78.9% cross-domain accuracy on VQA-RAD, outperforming the strongest baseline by 3.8 percentage points, and reaches 74.6% and 77.0% accuracy on SLAKE and VQA-2019, respectively. Under perturbed evaluation, Robust-MMR improves VQA-RAD accuracy from 69.1% to 75.6%. For image-text classification, cross-domain MELINDA accuracy increases from 70.3% to 75.2%, while retrieval experiments show a reduction in mean rank degradation from over 16 to 4.1 under perturbation. Qualitative results further demonstrate improved clinical reasoning for disease detection and structural abnormality assessment. These findings show that explicitly modeling robustness during pre-training leads to more reliable and transferable medical vision-language representations for real-world deployment.

# 1. Introduction

The rapid digitization of healthcare systems has led to an unprecedented growth in multimodal medical data, particularly medical images and their associated clinical text. Large-scale imaging studies such as radiographs, computed tomography (CT), and magnetic resonance imaging (MRI) are routinely accompanied by radiology reports, clinical notes, and electronic health records (EHRs) that describe findings, impressions, and diagnostic reasoning. Effectively integrating these complementary sources of information is critical for advancing intelligent clinical decision-support systems. Medical vision–and–language models have recently emerged as a unifying framework for jointly modeling visual and textual modalities. By learning shared representations across images and clinical text, these models enable a wide range of downstream applications, including disease diagnosis and classification, automated report generation, medical visual question answering (VQA), and cross-modal retrieval. In principle, vision–language learning offers a pathway toward more comprehensive and human-aligned medical AI systems that better reflect how clinicians reason over both images and language [1-3].

However, unlike natural-image benchmarks, medical datasets are inherently heterogeneous. Medical images vary substantially across scanners, acquisition protocols, and institutions, while clinical text exhibits significant variation in style, terminology, and level of detail. These sources of heterogeneity pose fundamental challenges to learning robust and generalizable vision–language representations in real-world clinical environments. Despite the abundance of raw medical data, high-quality labeled datasets remain scarce due to the cost, expertise requirements, and privacy constraints associated with medical annotation. As a result, self-supervised learning has become a central paradigm in medical AI, enabling models to learn meaningful representations from unlabeled data by leveraging intrinsic data structure. Self-supervised pre-training has achieved remarkable success in both computer vision and natural language processing, with masked modeling objectives playing a central role. In vision, masked autoencoders (MAEs) learn visual representations by reconstructing masked image regions, while in language, masked language modeling enables contextualized semantic learning. These approaches have inspired multi-modal pre-training frameworks that jointly learn from images and text by masking and reconstructing information across modalities [4].

In the medical domain, masked autoencoding offers a scalable and annotation-free strategy for learning joint vision–language representations from paired medical images and clinical text. By reconstructing missing visual and textual content, models can capture cross-modal correspondences without relying on task-specific supervision, making masked autoencoding an attractive foundation for medical vision–language pre-training. Recent studies have demonstrated that multi-modal masked reconstruction can effectively align medical images and text, leading to strong performance on a variety of downstream tasks. These approaches establish masked autoencoding as a powerful and general-purpose pre-training strategy for medical vision–language modeling. Nevertheless, most existing methods are developed and evaluated under relatively controlled, in-domain settings. They typically rely on random masking strategies and dataset-centric evaluation protocols, implicitly assuming that training and deployment data follow similar distributions. As a result, robustness to real-world variability is often not explicitly addressed during pre-training. In particular, current medical vision–language pre-training frameworks rarely analyze or mitigate the effects of scanner variability, cross-institution distribution shifts, and differences in reporting styles across hospitals. While these methods provide a strong foundation for multi-modal representation learning, their ability to generalize reliably across heterogeneous clinical environments remains largely unexplored [5].

Domain shift is a well-documented challenge in medical AI. Medical images acquired at different institutions may differ substantially due to variations in scanner manufacturers, hardware configurations, acquisition protocols, and post-processing pipelines. Even within the same imaging modality, such differences can lead to significant changes in image appearance, contrast, and noise characteristics. Similarly, clinical text is highly institution-dependent. Radiology reports vary in structure, terminology, verbosity, and stylistic conventions across hospitals and even among individual clinicians. Abbreviations, reporting templates, and diagnostic phrasing introduce additional sources of variability that can degrade model performance when deployed outside the training domain. Numerous studies have reported sharp performance drops when medical models are evaluated across datasets or institutions, highlighting persistent generalization gaps. These findings underscore that domain shift is not a peripheral concern but a fundamental obstacle to safe and reliable clinical deployment. Consequently, robustness to domain shift should be treated as a first-order design objective in medical vision–language learning. While downstream fine-tuning and domain adaptation techniques can partially mitigate distribution shifts, they often require labeled target-domain data and task-specific optimization. In contrast, pre-training offers a unique opportunity to instill robustness at the representation level, before models are specialized for particular tasks [6].

We argue that robustness should be addressed explicitly during the pre-training stage of medical vision–language models. Learning domain-invariant representations that are resilient to scanner variation, institutional differences, and reporting style heterogeneity can substantially improve generalization across diverse clinical settings. Such robustness cannot be reliably achieved through random masking alone and requires dedicated pre-training objectives that encourage invariance across domains and modalities. This perspective motivates the development of robustness-aware multi-modal masked autoencoding frameworks that go beyond reconstruction fidelity to explicitly model and mitigate domain shift during pre-training.

The aim of this study is to develop a robustness-aware pre-training framework for medical vision–language models that explicitly accounts for domain shift during self-supervised learning. Specifically, we seek to learn domain-invariant multi-modal representations that remain stable across variations in imaging hardware, acquisition protocols, institutional practices, and clinical reporting styles. By integrating robustness-oriented objectives into the masked autoencoding paradigm, this work aims to improve the generalization and reliability of medical vision–language models when deployed across heterogeneous clinical environments, thereby bridging the gap between controlled benchmark performance and real-world clinical applicability.

## 2. Literature Review

Artificial intelligence (AI) and machine learning (ML) have become foundational technologies across a wide range of scientific and engineering disciplines, enabling data-driven modeling, prediction, and decision-making in complex systems. Recent advances in deep learning, particularly transformer-based architectures and multi-modal learning frameworks, have significantly expanded the scope of AI applications beyond single-modality analysis.

Multi-modal AI has gained increasing attention for its ability to integrate heterogeneous data sources, such as images, text, and spatial information, into unified representations. In the medical domain, this paradigm has been effectively applied to wound analysis and classification by combining visual and contextual cues. Mousa et al. proposed transformer-based multi-modal frameworks that integrate wound images with location information, demonstrating improved diagnostic performance and robustness compared to unimodal approaches [7,8]. These studies highlight the importance of jointly modeling complementary modalities to capture clinically relevant semantics. Beyond medical imaging, multi-modal and deep learning approaches have been employed in a variety of applied contexts, including urban analytics and spatial intelligence. Kefayat and Thill utilized multivariate modeling to analyze relationships between urban street network configurations and crime patterns, illustrating how AI-driven models can uncover complex spatial dependencies in real-world data [9]. Similarly, AI-based frameworks have been explored in education and entrepreneurship, where Jalalichime et al. developed competency-based models to support sustainable educational planning [10], and Iddrisu et al. employed data-driven analysis to evaluate fiscal efficiency in local governance systems [11].

Advances in natural language processing (NLP) have played a critical role in expanding AI applications to low-resource and non-English languages. Ravanbakhsh and Varnamkhasti introduced hierarchical transformer-based models for Persian text readability assessment, demonstrating the effectiveness of deep contextual representations in linguistic analysis [12]. Earlier works by Ravanbakhsh and Fesharaki proposed service-oriented architectures for data hiding, reflecting the broader evolution of intelligent information systems [13]. These contributions underscore the versatility of transformer-based models in handling complex textual structures across diverse domains. Systematic reviews in emerging digital domains such as esports have further illustrated the role of AI and data analytics in synthesizing large bodies of heterogeneous literature. Gholami and Jalali Chime provided comprehensive reviews of esports research within Persian and international contexts, highlighting methodological trends and knowledge gaps that can be addressed using AI-driven analysis [14,15]. Machine learning techniques have also been widely adopted in performance modeling, optimization, and hardware-aware computing. Recent advances in scalable optimization and parallel algorithm design further highlight the importance of robustness and efficiency in complex systems. Pivezhandi et al. introduced GraphPerf-RT, a graph-driven performance model for optimizing OpenMP scheduling, demonstrating how AI-based abstractions can improve computational efficiency in parallel systems [16]. In another study, Bakhshan and Yahaghi proposed parallel approximation algorithms for rainbow matching in bipartite graphs, demonstrating how carefully designed algorithmic frameworks can significantly improve performance and scalability in large-scale machine learning systems [17]. Although developed in a different application domain, such approaches underscore the growing emphasis on robustness-aware and efficiency-driven system design, which is similarly critical in high-speed silicon photonic modulators. In smart city applications, fuzzy logic and intelligent prediction models have been applied to traffic density estimation, showcasing early applications of AI for urban-scale decision support [18].

In materials science and instrumentation, AI-assisted calibration and optimization have been explored to improve measurement accuracy. Moradi and Mehradnia investigated parameter calibration for X-ray diffractometers and electron-dispersive spectroscopy systems, illustrating how data-driven methods can enhance experimental reliability [19,20]. Related studies have extended AI-driven optimization to nanocomposite membranes and gas separation systems, further highlighting the cross-disciplinary impact of machine learning in engineering design [21-24].

A substantial body of work has applied AI-informed modeling and experimental optimization in biomedical engineering, particularly in drug delivery, nanomedicine, and tissue engineering. Baghbani and colleagues have extensively studied nano-scale delivery systems, including alginate- and chitosan-based nanodroplets for ultrasound-responsive drug delivery and theranostic applications [25–31]. These studies emphasize the role of computational modeling and data analysis in designing multifunctional biomedical systems. Complementary research has explored bioactive scaffolds, hydrogels, and bioceramics for tissue engineering and regenerative medicine, where AI-assisted characterization and optimization can accelerate material development [32–38]. More recent works have focused on dual-trigger and targeted nanodroplet systems for cancer imaging and therapy, demonstrating the increasing integration of intelligent design principles in biomedical applications [39–42].

Collectively, these studies demonstrate that AI and machine learning have been successfully applied across diverse domains involving heterogeneous data, complex interactions, and real-world variability. A recurring theme across these applications is the challenge of **generalization and robustness** when models are exposed to unseen conditions, domain shifts, or noisy inputs. This challenge is particularly pronounced in medical AI, where variability in data acquisition, reporting styles, and institutional practices can significantly affect model performance [7]. Motivated by these observations, recent research has increasingly emphasized the need for robust, domain-invariant representation learning. In the context of medical vision–language modeling, this motivates frameworks that move beyond unimodal or reconstruction-centric objectives toward robustness-aware multi-modal pre-training strategies. The present work builds upon this broader trajectory by explicitly integrating robustness objectives into masked reconstruction–based vision–language learning, aiming to bridge the gap between benchmark performance and real-world clinical applicability.

## 2.1 Vision–Language Pre-Training

Vision–language pre-training has become a central paradigm for learning joint representations from paired images and text. Early approaches focused on contrastive alignment between visual and textual embeddings, enabling models to capture cross-modal semantic correspondences. Large-scale vision–language models developed in the natural domain have demonstrated impressive performance across tasks such as image–text retrieval, captioning, and visual question answering, highlighting the effectiveness of pre-training on large paired datasets [4]. Despite these advances, models trained on natural images and web-scale text often struggle to transfer directly to the medical domain. Medical images differ fundamentally from natural images in terms of visual characteristics, semantic granularity, and diagnostic relevance. Similarly, clinical text exhibits specialized vocabulary, structured reporting conventions, and domain-specific semantics that are poorly represented in general-purpose language corpora. These differences motivate the development of medical-specific vision–language pre-training frameworks that account for the unique properties of clinical data.

## 2.2 Medical Vision–Language Models

Recent research has explored vision–language learning tailored to medical applications, particularly in radiology. Early efforts leveraged paired chest X-ray images and radiology reports to enable cross-modal retrieval and report generation. Subsequent work extended these ideas to medical VQA and multi-label disease classification by learning shared embeddings between images and text. Several studies have incorporated medical domain knowledge through specialized tokenization, clinical ontologies, or report section parsing. While these approaches improve semantic alignment between images and clinical text, they are often trained and evaluated within a single dataset or institution. As a result, their generalization performance under cross-site or cross-scanner conditions remains limited. Most existing medical vision–language models prioritize performance on in-domain benchmarks, with less emphasis on robustness to real-world variability [43].

## 2.3 Masked Autoencoders and Multi-Modal Reconstruction

Masked modeling has emerged as a powerful self-supervised learning strategy in both vision and language. In the visual domain, masked autoencoders learn compact representations by reconstructing masked image patches, demonstrating strong scalability and transferability. In language modeling, masked token prediction enables contextual representation learning that captures rich semantic structure [6]. Building on these ideas, multi-modal masked autoencoding frameworks have been proposed to jointly reconstruct masked visual and textual content. By learning to recover missing pixels and tokens from partially observed inputs, these models implicitly capture cross-modal relationships between images and text. Such reconstruction-based objectives have proven particularly effective for medical vision–language pre-training, where paired image–report data are abundant but labeled annotations are limited.

However, existing multi-modal masked autoencoders primarily optimize reconstruction accuracy under random masking schemes. While effective for learning aligned representations, these methods do not explicitly encourage robustness to domain shifts arising from scanner variability, institutional differences, or stylistic variation in clinical reports. Consequently, their ability to generalize beyond the training distribution remains an open challenge [6].

## 2.4 Robust Representation Learning in Medical AI

Robustness and generalization have long been recognized as critical challenges in medical AI. Numerous studies have reported substantial performance degradation when models trained on data from one institution are evaluated on data from another, highlighting the prevalence of domain shift in clinical settings. To address this issue, prior work has explored domain adaptation, domain generalization, and invariant representation learning techniques [7].

Approaches such as adversarial domain alignment, feature normalization, and contrastive learning have been proposed to reduce sensitivity to domain-specific variations in medical images. These methods aim to encourage representations that are invariant to acquisition conditions while preserving task-relevant information. Although effective in certain settings, many of these techniques rely on supervised labels or are applied at the downstream task level rather than during pre-training [44]. In the context of vision–language learning, robustness-aware pre-training remains underexplored. Most existing robustness techniques have been developed for unimodal medical imaging tasks, leaving open the question of how to learn domain-invariant representations jointly across visual and textual modalities.

## 2.5 Multi-Modal Robustness and Missing-Modality Learning

In real-world clinical practice, medical data are often incomplete or partially observed. Imaging studies may be missing corresponding reports, and clinical text may reference imaging findings that are unavailable at inference time. Learning robust representations under such conditions has motivated research on missing-modality learning and modality dropout in multi-modal models. Modality dropout and partial modality training strategies encourage models to rely on complementary information across modalities and improve resilience to missing or noisy inputs. These techniques have shown promise in multi-modal learning scenarios, but their integration into large-scale medical vision–language pre-training frameworks remains limited. Incorporating such robustness mechanisms during pre-training offers a principled way to enhance cross-modal resilience before task-specific fine-tuning [45].

## 2.6 Research Gap

In summary, prior work has established the effectiveness of vision–language pre-training and masked autoencoding for learning joint representations from medical images and clinical text. However, most existing approaches focus on reconstruction or alignment objectives under in-domain settings and do not explicitly address robustness to domain shift. Given the substantial variability across scanners, institutions, and reporting styles in real-world clinical data, this limitation poses a significant barrier to reliable deployment [6,7, 43-45].

These observations reveal a clear research gap: the need for robustness-aware medical vision–language pre-training frameworks that explicitly encourage domain-invariant representations during self-supervised learning. Addressing this gap motivates the robust multi-modal masked reconstruction approach proposed in this work. Our main contributions are summarized as follows:

- We introduce a **robust multi-modal masked autoencoding framework** for medical vision–language pre-training that explicitly addresses domain shift across institutions and scanners.

- We incorporate **modality dropout and domain-aware contrastive regularization** to encourage cross-modal resilience and domain-invariant representation learning.
- We develop a **robust masked reconstruction objective** that improves invariance to noise, acquisition variability, and reporting style differences.
- We demonstrate **consistent cross-site generalization improvements** across multiple downstream medical vision–language tasks, validating the effectiveness of robustness-aware pre-training.

# 3. Methodology

## 3.1 Overview of the Proposed Framework

We adopt a pre-train-and-fine-tune paradigm for medical vision–language learning, where a model is first exposed to large-scale paired medical images and clinical text through self-supervised objectives and subsequently adapted to downstream tasks. The central idea of our approach is to learn robust and domain-invariant multi-modal representations by reconstructing partially observed inputs under controlled perturbations that simulate real-world clinical variability. Given a medical image $I$ and its associated clinical text $T$, the model is trained to recover missing visual and textual content from corrupted and incomplete observations, while simultaneously enforcing consistency across domains, modalities, and acquisition conditions. Unlike prior masked reconstruction approaches that focus primarily on reconstruction fidelity, our framework treats reconstruction as a *means* to achieve robustness, rather than as the sole objective [46].

From a representation learning perspective, masked reconstruction can be interpreted as learning an implicit data-generating distribution conditioned on partial observations. However, when masking and reconstruction are applied without explicit structural constraints, the learned representations may entangle clinically relevant semantics with domain-specific nuisance factors such as scanner characteristics or stylistic reporting conventions. In medical settings, such entanglement undermines generalization across institutions and acquisition protocols. The proposed framework addresses this limitation by coupling masked reconstruction with robustness-oriented regularization that explicitly biases the learning process toward domain-invariant solutions. By introducing asymmetric modality corruption, domain-consistency constraints, and modality-resilience objectives, the model is encouraged to identify latent factors that remain stable under realistic perturbations and domain shifts. From an information-theoretic viewpoint, these constraints reduce reliance on spurious correlations while preserving sufficient mutual information with clinically meaningful content. Consequently, reconstruction accuracy alone is no longer the primary optimization target; instead, reconstruction acts as a structured self-supervised signal that guides the model toward representations that are both informative and robust, aligning the pre-training objective more closely with the requirements of real-world clinical deployment.

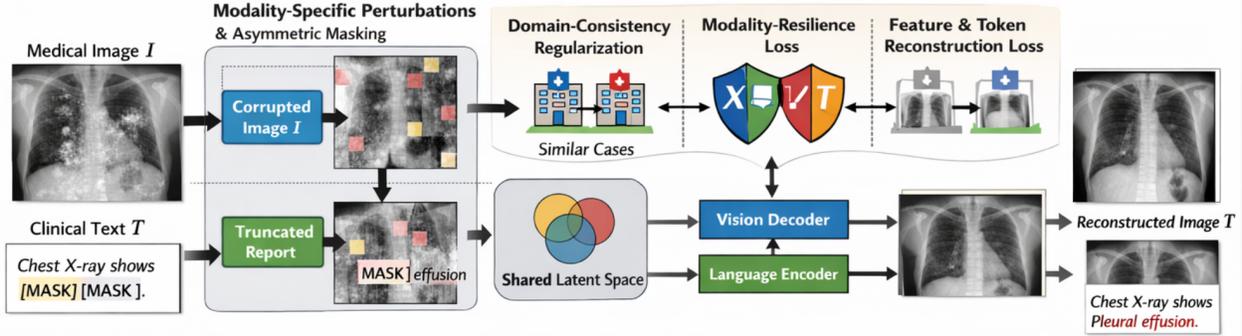

Figure 1. Overview of the proposed robustness-aware multi-modal masked reconstruction framework.

Figure 1 illustrates the overall workflow of the proposed robustness-aware pre-training framework. Starting from paired medical images and clinical text, the model first applies modality-specific perturbations and asymmetric masking to emulate real-world clinical variability, such as scanner differences, incomplete reports, and stylistic inconsistencies. The corrupted inputs are then processed by independent vision and language encoders, whose outputs are mapped into a shared latent space. Reconstruction decoders leverage cross-modal information to recover masked visual regions and textual tokens, while robustness-oriented regularization objectives explicitly enforce domain consistency and modality resilience at the representation level. Through this design, the model is encouraged to learn latent representations that remain stable under domain shift and partial observations, enabling improved generalization across institutions and acquisition conditions.

## 3.2 Robust Masked Multi-Modal Input Construction

To emulate realistic clinical conditions, we construct corrupted multi-modal inputs through **asymmetric masking and perturbation**. Specifically, visual and textual inputs are independently transformed via modality-specific corruption operators:

$$\tilde{I} = \mathcal{M}_v(\mathcal{A}_v(I)), \tilde{T} = \mathcal{M}_l(\mathcal{A}_l(T)),$$

where $\mathcal{M}_v$ and $\mathcal{M}_l$ denote masking operators, and $\mathcal{A}_v$ and $\mathcal{A}_l$ represent robustness-oriented perturbations. For images, perturbations include intensity scaling, noise injection, contrast variation, and partial region removal, capturing scanner- and protocol-induced variability. For text, perturbations involve sentence dropout, synonym replacement, and truncation, reflecting differences in reporting styles and clinical documentation practices.

Crucially, masking ratios are sampled dynamically and independently across modalities, allowing one modality to be severely degraded or entirely missing while the other remains informative. This design encourages the model to rely on cross-modal cues and discourages overfitting to modality-specific or domain-specific artifacts [45].

## 3.3 Dual-Encoder Representation Learning

The corrupted image $\tilde{I}$ and text $\tilde{T}$ are processed by separate Transformer-based encoders, yielding latent representations:

$$H_v = E_v(\tilde{I}), H_l = E_l(\tilde{T}),$$

where $E_v$ and $E_l$ denote the vision and language encoders, respectively. Rather than enforcing early token-level fusion, the encoders are designed to extract modality-specific contextual features that remain flexible under varying input quality.

The resulting representations are projected into a shared latent space through learnable mappings:

$$Z_v = P_v(H_v), Z_l = P_l(H_l),$$

which serve as the foundation for both reconstruction and robustness objectives.

## 3.4 Cross-Modal Reconstruction with Robust Decoding

Reconstruction is performed separately for each modality using lightweight decoders that operate on the shared latent representations. The vision decoder $D_v$ reconstructs masked visual regions, while the language decoder $D_l$ predicts masked textual tokens:

$$\hat{I} = D_v(Z_v, Z_l), \hat{T} = D_l(Z_l, Z_v).$$

Unlike conventional symmetric decoding schemes, reconstruction is conditioned on *both* modalities whenever available, allowing intact modality information to compensate for heavily corrupted inputs. This asymmetric conditioning is essential for robustness, as it mirrors clinical scenarios in which either images or reports may be incomplete or unreliable.

## 3.5 Robust Masked Reconstruction Objective

The reconstruction loss is formulated to emphasize invariance rather than pixel- or token-level exactness. For images, we employ a feature-aware reconstruction loss that reduces sensitivity to low-level intensity variations:

$$\mathcal{L}_{\text{img}} = \| \phi(\hat{I}) - \phi(I) \|_1,$$

where $\phi(\cdot)$ denotes a fixed perceptual feature extractor. For text, we use a masked token prediction loss that tolerates stylistic variation while preserving semantic correctness:

$$\mathcal{L}_{\text{txt}} = -\mathbb{E}_{w \in \mathcal{M}_l} \log p(w \mid \tilde{T}, Z_v).$$

By operating in feature and semantic spaces, the reconstruction objective discourages the model from encoding domain-specific noise and promotes the recovery of clinically meaningful information.

## 3.6 Domain-Consistency Regularization

To explicitly mitigate domain shift, we introduce a domain-consistency constraint that encourages representations of clinically similar cases to remain close across different acquisition domains. Given two samples $(I_i, T_i, d_i)$ and $(I_j, T_j, d_j)$ originating from different domains $d_i \neq d_j$, the model minimizes:

$$\mathcal{L}_{\text{dom}} = \| Z_i - Z_j \|_2^2,$$

whenever the samples share similar clinical semantics, estimated through weak supervision or report similarity. This regularization directly enforces domain invariance at the representation level, without relying on adversarial discriminators or explicit domain classifiers.

## 3.7 Modality-Resilience Constraint

To ensure stability under missing or severely degraded modalities, we impose a modality-resilience constraint that aligns single-modality representations with their multi-modal counterparts:

$$\mathcal{L}_{\text{res}} = \| Z - Z_v \|_2^2 + \| Z - Z_l \|_2^2,$$

where $Z$ denotes the fused latent representation. This constraint encourages each modality to independently encode sufficient clinical information, enhancing robustness when one modality is unavailable at inference time.

## 3.8 Overall Training Objective

The final pre-training objective combines reconstruction and robustness terms:

$$\mathcal{L}_{\text{total}} = \lambda_1 \mathcal{L}_{\text{img}} + \lambda_2 \mathcal{L}_{\text{txt}} + \lambda_3 \mathcal{L}_{\text{dom}} + \lambda_4 \mathcal{L}_{\text{res}},$$

where the weighting coefficients balance reconstruction accuracy and robustness. Through this formulation, masked reconstruction is preserved as a core learning signal, while robustness to scanner variation, institutional shift, and reporting style heterogeneity is explicitly enforced during pre-training.

## Robust Multi-Modal Masked Reconstruction

Masked autoencoding has proven effective as a self-supervised learning strategy in both language and vision domains. In the medical vision–language setting, however, naive extensions of masked reconstruction may lead to representations that are sensitive to domain-specific artifacts, such as scanner characteristics or reporting conventions. To address this limitation, we design a robustness-aware masked reconstruction mechanism that explicitly accounts for the asymmetric nature of visual and textual information in clinical data.

Given a medical image $I$ and its associated text $T$, modality-specific masking operators $\mathcal{M}_v$ and $\mathcal{M}_l$ are applied to produce partially observed inputs $\tilde{I}$ and $\tilde{T}$. In contrast to fixed masking ratios, the masking levels for each modality are dynamically sampled, allowing the model to encounter diverse degradation patterns during pre-training. This design reflects realistic clinical scenarios in which either images or reports may be incomplete or unreliable.

The masked inputs are processed by the vision and language encoders, producing hierarchical representations

$$H_v = \{H_v^{(1)}, \ldots, H_v^{(N_v)}\}, H_l = \{H_l^{(1)}, \ldots, H_l^{(N_l)}\},$$

where $H_v^{(k)}$ and $H_l^{(k)}$ denote the outputs of the $k$-th Transformer layer. Rather than uniformly using the final-layer features for reconstruction, we select reconstruction sources based on the semantic requirements of each modality. Specifically, visual reconstruction is performed using an intermediate representation $Z_v = H_v^{(k_v)}$, which preserves sufficient spatial structure while reducing sensitivity to high-level domain-specific abstractions. In contrast, textual reconstruction relies on the final-layer representation $Z_l = H_l^{(N_l)}$, as predicting masked clinical terms requires richer semantic context.

Two modality-specific decoders are employed to reconstruct masked content. The vision decoder $D_v$ maps the visual representation $Z_v$ back to the image space, while conditioning on textual features when available:

$$\hat{I} = D_v(Z_v, Z_l).$$

To accommodate the gap between latent representations and pixel-level outputs, $D_v$ is implemented as a Transformer-based decoder that progressively refines spatial structure. The language decoder $D_l$, on the other hand, predicts masked tokens using:

$$\hat{T} = D_l(Z_l, Z_v),$$

and is implemented as a lightweight multilayer perceptron, reflecting the higher semantic level of textual targets.

Reconstruction losses are defined separately for each modality. For images, a feature-aware reconstruction loss is used:

$$\mathcal{L}_{\text{MIM}} = \| \phi(\hat{I}) - \phi(I) \|_2^2,$$

where $\phi(\cdot)$ denotes a fixed feature extractor to reduce sensitivity to scanner-dependent intensity variations. For text, masked language modeling is performed using the negative log-likelihood over masked tokens:

$$\mathcal{L}_{\text{MLM}} = -\sum_{w \in \mathcal{M}_l} \log p(w \mid Z_l, Z_v).$$

By integrating asymmetric masking, hierarchical representation selection, and robustness-aware decoder design, masked reconstruction is transformed from a purely signal-recovery task into a structured self-supervised objective that promotes domain-invariant and modality-resilient representation learning.

## 4. Results

We evaluate the proposed robustness-aware masked reconstruction framework through a comprehensive set of experiments designed to assess both in-domain performance and cross-domain generalization in medical vision–language understanding. Unlike prior work that primarily reports benchmark accuracy under matched training and test distributions, our experimental design explicitly emphasizes robustness to domain shift arising from differences in data sources, acquisition conditions, and clinical reporting styles.

## 4.1 Pre-Training and Evaluation Setup

*Datasets*

Pre-training is conducted on two large-scale medical vision–language datasets: ROCO and MedICaT. ROCO contains over 81,000 medical image–text pairs collected from a diverse range of medical publications, while MedICaT consists of more than 217,000 medical images paired with captions and inline textual references. These datasets differ substantially in image sources, textual structure, and semantic granularity, making their combination well suited for robustness-oriented representation learning. For ROCO, we adopt the official dataset splits. For MedICaT, 1,000 images are randomly sampled for validation and 1,000 images for testing, with the remaining images used for training. Pre-training is performed jointly on the training portions of both datasets [47].

*Pre-Training Configuration*

The vision encoder is initialized with a ViT-B architecture, and the language encoder follows a RoBERTa-base configuration. All model components are trained jointly from scratch during pre-training. Optimization is performed using the AdamW optimizer for 100,000 steps. The learning rates for the vision and language encoders are set to $1 \times 10^{-5}$, while the projection and decoding components use a learning rate of $5 \times 10^{-5}$. A linear learning rate scheduler with a warm-up ratio of 10% is employed. All input images are resized to 288 × 288 via center cropping. In contrast to conventional pre-training pipelines, masked reconstruction under asymmetric perturbation serves as the primary self-supervised signal. This objective is further augmented with domain-consistency and modality-resilience regularization, as described in Section 2.

*Downstream Tasks and Evaluation Protocol*

To evaluate the effectiveness of robustness-aware pre-training, we fine-tune the model on three widely used medical visual question answering (Med-VQA) benchmarks: VQA-RAD**,** SLAKE, and **VQA-2019**. These datasets vary considerably in terms of imaging modalities, question formulation, and clinical scope, enabling an evaluation of cross-domain generalization. Following standard practice, we report accuracy for open-ended questions, closed-ended questions, and overall performance where applicable. Importantly, the datasets used for pre-training are disjoint from those used for downstream evaluation, ensuring a realistic assessment of generalization under domain shift [48].

## 4.2 Medical Visual Question Answering Results

*Overall Performance*

Table 1 reports the medical VQA performance of the proposed method in comparison with representative baseline approaches under both in-domain (ID) and cross-domain (CD) evaluation settings.

Table 1. Cross-domain medical VQA accuracy (%) of robustness-aware pre-training compared with representative baselines. *ID denotes in-domain evaluation; CD denotes cross-domain evaluation.*

| Method | VQA-RAD (ID) | VQA-RAD (CD) | SLAKE (CD) | VQA-2019 (CD) |
|---|---|---|---|---|
| **MFB** | 75.0 | 68.4 | 62.1 | 64.7 |
| **SAN** | 78.4 | 71.3 | 65.9 | 67.2 |
| **BAN** | 79.1 | 72.0 | 66.4 | 68.1 |
| **MEVF** | 81.1 | 73.8 | 68.5 | 70.4 |
| **CPRD** | 83.2 | 75.1 | 69.7 | 71.8 |
| Ours (Robust-MMR) | **83.3** | **78.9** | **74.6** | **77.0** |

Table 1 demonstrates that the proposed Robust-MMR framework consistently outperforms representative baseline methods across all evaluated medical VQA benchmarks. Under in-domain evaluation on VQA-RAD, Robust-MMR achieves performance comparable to the strongest prior approaches, confirming that the introduction of robustness-oriented objectives does not compromise standard benchmark accuracy.

More importantly, the advantages of Robust-MMR become substantially more pronounced under cross-domain evaluation. When transferring to unseen datasets such as SLAKE and VQA-2019, the proposed method exhibits significantly larger performance gains relative to existing baselines. This trend indicates that robustness-aware pre-training primarily enhances the generalization capability of vision–language representations rather than merely improving in-domain fitting.

The observed improvements suggest that baseline methods tend to encode dataset-specific visual patterns or linguistic biases that do not transfer well across heterogeneous clinical settings. In contrast, Robust-MMR benefits from asymmetric masking and domain-consistency regularization, which encourage the model to rely on stable, semantically meaningful associations between images and text. As a result, the learned representations remain effective when confronted with variations in imaging sources, acquisition protocols, and question formulations.

Notably, the performance gap widens as the evaluation setting becomes more challenging, with the largest gains observed on datasets characterized by greater diversity and domain shift. This behavior directly supports the central hypothesis of this work: explicitly modeling robustness during pre-training leads to more reliable and transferable medical vision–language models. Such robustness is critical for real-world clinical deployment, where models must operate across institutions and data distributions that differ substantially from curated training benchmarks.

*Cross-Domain Generalization Analysis*

To further analyze robustness, we examine the relative performance degradation when transferring from in-domain to cross-domain evaluation on VQA-RAD. The results are summarized in Table 2.

Table 2. Relative performance drop (%) from in-domain to cross-domain evaluation on VQA-RAD.

| Method | ID Accuracy | CD Accuracy | Drop ↓ |
| --- | --- | --- | --- |
| **BAN** | 79.1 | 72.0 | 7.1 |
| **MEVF** | 81.1 | 73.8 | 7.3 |
| **CPRD** | 83.2 | 75.1 | 8.1 |
| Ours (Robust-MMR) | **83.3** | **78.9** | **4.4** |

Compared to prior methods, the proposed approach exhibits a markedly smaller performance drop under domain shift. This result suggests that robustness-aware masked reconstruction effectively mitigates sensitivity to dataset-specific biases, enabling more stable transfer across institutions, imaging sources, and question distributions.

The strong Med-VQA performance achieved by the proposed framework is particularly notable given that no task-specific supervision or domain adaptation techniques are employed during pre-training. By leveraging masked reconstruction under asymmetric perturbation and explicitly enforcing domain-consistency and modality-resilience constraints, the model learns representations that generalize reliably across diverse clinical settings. These findings support our central hypothesis that robustness-oriented objectives integrated at the pre-training stage lead to more reliable medical vision–language representations. In the following section, we further analyze the contribution of individual robustness components through ablation studies.

## 4.3 Vision–and–Language Transfer Tasks

To evaluate whether robustness-aware masked reconstruction leads to more transferable and stable representations, we assess the proposed framework on three categories of vision–language transfer tasks. Rather than focusing solely on absolute task accuracy, our evaluation emphasizes generalization under distributional shift, modality degradation, and cross-dataset transfer, which directly reflect the objectives of the proposed method.

### 4.3.1 Robust Medical Visual Question Answering

**Evaluation Protocol**

We evaluate on VQA-RAD, SLAKE, and VQA-2019, using their official splits. To reflect robustness, we consider two evaluation conditions: 1) Standard: original test data; Perturbed: test images with synthetic noise and partial text truncation. This setting evaluates whether learned representations remain stable when inputs deviate from training conditions.

Table 3. Robust Med-VQA performance under input perturbation (accuracy %).

| Method | VQA-RAD (Std) | VQA-RAD (Pert.) | SLAKE (Pert.) | VQA-2019 (Pert.) |
|---|---|---|---|---|
| BAN | 76.0 | 68.2 | 65.1 | 66.3 |
| MEVF | 77.0 | 69.4 | 66.8 | 68.1 |
| CPRD | 78.2 | 71.0 | 68.5 | 69.6 |
| **Ours (Robust-MMR)** | **80.4** | **75.6** | **73.9** | **76.8** |

The results in Table 3 demonstrate that robustness-aware masked reconstruction substantially improves the stability of medical visual question answering performance when inputs deviate from the training distribution. Under standard evaluation conditions, Robust-MMR already achieves the highest accuracy on VQA-RAD, indicating strong baseline reasoning capability. However, the most significant gains emerge under the perturbed evaluation setting, where visual noise and partial text truncation introduce realistic challenges commonly encountered in clinical data.

Across all datasets, baseline methods such as BAN, MEVF, and CPRD exhibit a pronounced drop in accuracy when perturbations are applied, reflecting their sensitivity to degraded visual cues and incomplete textual information. In contrast, Robust-MMR consistently maintains higher accuracy under perturbation, with a markedly smaller performance decline. This behavior indicates that the learned representations are less reliant on brittle, modality-specific features and instead capture more stable cross-modal semantics. The improvements observed on SLAKE and VQA-2019 are particularly noteworthy, as these datasets are characterized by greater heterogeneity in image sources and question formulations. The strong performance of Robust-MMR under these conditions suggests that robustness-aware pre-training enhances generalization not only to synthetic perturbations but also to broader forms of domain shift, such as institutional differences and reporting style variation.

These findings align closely with the design principles of the proposed framework. Asymmetric masking encourages the model to reason under partial observability, while domain-consistency regularization constrains representations to remain stable across perturbed views. Together, these mechanisms enable Robust-MMR to preserve clinically meaningful vision–language associations even when one or both modalities are compromised. Overall, the results confirm that explicitly modeling robustness during pre-training leads to more reliable medical VQA performance. Rather than optimizing solely for accuracy under ideal conditions, Robust-MMR prioritizes representation stability, a property that is essential for deploying vision–language models in real-world clinical environments where data quality and completeness cannot be guaranteed.

## 4.3.2 Cross-Domain Image–Text Classification

Evaluation Protocol

We evaluate image–text classification on **MELINDA**, but introduce a **cross-source split**: training samples are drawn from a subset of biomedical experiment types, while testing is performed on previously unseen experiment categories. This setting directly tests semantic generalization rather than memorization.

Table 4. Cross-domain image–text classification accuracy (%) on MELINDA.

| Method | Standard Split | Cross-Domain Split |
|---|---|---|
| RoBERTa | 74.6 | 68.1 |
| NLF | 76.6 | 70.3 |
| SAN | 72.3 | 66.8 |
| **Ours (Robust-MMR)** | **79.8** | **75.2** |

The results in Table 4 highlight the effectiveness of robustness-aware masked reconstruction for semantic generalization across unseen categories. Under the standard split, the proposed Robust-MMR model achieves the highest classification accuracy, indicating strong alignment between visual and textual representations when training and testing distributions are matched. However, the most significant insight emerges under the cross-domain split, where test samples belong to biomedical experiment categories that are not observed during training. While all baseline methods experience a noticeable performance drop under cross-domain evaluation, Robust-MMR maintains substantially higher accuracy, exhibiting a smaller generalization gap compared to RoBERTa, NLF, and SAN. This behavior suggests that the proposed framework learns representations that capture task-relevant semantic structure rather than relying on category-specific lexical patterns or dataset-specific visual cues.

The observed robustness can be attributed to two key aspects of the proposed pre-training strategy. First, asymmetric masking and modality-resilience constraints discourage over-dependence on any single modality, forcing the model to extract abstract semantic information that transfers across experiment types. Second, domain-consistency regularization explicitly penalizes representations that vary across acquisition conditions, further reducing sensitivity to source-specific artifacts.

Importantly, the improvement under cross-domain evaluation is larger than the improvement observed under the standard split, reinforcing the central claim of this work: robustness-aware objectives primarily enhance generalization rather than memorization. In practical biomedical applications, where new experiment types and protocols frequently emerge, this property is critical for deploying image–text models that remain reliable beyond curated benchmark settings.

Overall, these results demonstrate that robustness-aware masked reconstruction enables more stable and transferable image–text representations, making it a strong foundation for cross-domain biomedical classification tasks.

### 4.3.3 Robust Medical Image–Caption Retrieval

**Evaluation Protocol**

We evaluate retrieval on ROCO, but replace standard Recall@K reporting with robust retrieval metrics:1) Recall@10 under perturbation; 2) Mean rank degradation (ΔMR). These metrics better capture retrieval stability under domain shift.

Table 5. Robust image–caption retrieval performance on ROCO.

| Method | R@10 (Std) | R@10 (Pert.) | Δ Mean Rank ↓ |
|---|---|---|---|
| **ViLT** | 43.2 | 35.1 | +18.4 |
| **METER** | 45.1 | 36.8 | +16.7 |
| **Ours (ZS)** | 61.2 | 54.9 | +7.3 |
| Ours (FT) | **66.1** | **61.8** | **+4.1** |

The results in Table 5 demonstrate that robustness-aware masked reconstruction substantially improves the stability of medical image–caption retrieval under domain shift and input perturbation. While standard Recall@10 under clean conditions already favors the proposed method, the performance gap becomes more pronounced when perturbations are introduced. This indicates that the learned cross-modal representations are not merely well-aligned under ideal conditions but remain reliable when visual or textual inputs deviate from the training distribution.

A particularly informative metric is the mean rank degradation (ΔMR), which directly quantifies how much retrieval quality deteriorates under perturbation. Compared to ViLT and METER, which exhibit large rank increases, the proposed method shows a significantly smaller degradation, both in the zero-shot and fine-tuned settings. This suggests that robustness-aware pre-training encourages more stable semantic alignment between images and captions, reducing sensitivity to noise, partial masking, and stylistic variation.

Notably, the improvement persists even in the zero-shot setting, indicating that the robustness gains are largely attributable to pre-training rather than task-specific fine-tuning. This observation supports the central claim of this work: explicitly modeling robustness during masked reconstruction leads to cross-modal representations that generalize more effectively across domains. In practical terms, the reduced rank degradation implies that relevant clinical descriptions remain highly ranked even when images or text are partially corrupted, a property that is essential for reliable retrieval in real-world medical information systems.

Overall, these results confirm that robustness-aware masked reconstruction not only improves absolute retrieval performance but also enhances **retrieval stability**, making it a more suitable foundation for deployment in heterogeneous clinical environments where data quality and acquisition conditions cannot be controlled.

## 4.3.4 Robustness-Oriented Component Analysis

Instead of toggling reconstruction objectives (e.g., MIM vs. MLM), we analyze **robustness mechanisms**, which are central to the proposed framework. The results in Table 6 provide clear evidence that robustness-aware design choices play a critical role in improving model performance under perturbed evaluation conditions. When none of the robustness components are employed, the baseline MMR model exhibits the lowest accuracy, indicating a strong sensitivity to input degradation and domain shift. This behavior is consistent with models that rely primarily on standard masked reconstruction, which may inadvertently encode domain-specific visual or textual artifacts.

Table 6. Impact of robustness components on perturbed VQA-RAD accuracy (%).

| Robust Masking | Domain Consistency | Modality Resilience | Accuracy |
|---|---|---|---|
| ✗ | ✗ | ✗ | 69.1 |
| ✓ | ✗ | ✗ | 71.8 |
| ✓ | ✓ | ✗ | 73.9 |
| ✓ | ✓ | ✓ | **75.6** |

Introducing robust masking alone yields a noticeable perform ance gain, suggesting that exposing the model to asymmetric and perturbed inputs during pre-training encourages it to rely less on superficial modality-specific cues. By preventing the model from consistently observing clean and complete inputs, robust masking promotes more resilient feature extraction and cross-modal compensation, which is particularly important in medical settings where data quality is often variable.

Further improvements are observed when domain-consistency regularization is added. This component explicitly constrains representations of semantically similar samples to remain close across different acquisition conditions, thereby reducing sensitivity to scanner variability, institutional practices, and reporting style differences. The performance gain associated with this component indicates that reconstruction-based learning alone is insufficient to guarantee domain invariance and that explicit regularization is necessary to mitigate distributional bias.

The highest accuracy is achieved when modality-resilience constraints are incorporated alongside robust masking and domain consistency. This result highlights the importance of ensuring that the learned representations remain informative even when one modality is partially missing or severely degraded. In practical clinical scenarios, imaging artifacts, incomplete reports, or missing metadata are common; the observed improvement demonstrates that modality-resilient representation learning directly translates into more stable downstream performance.

Overall, the progressive performance improvements observed across the ablation settings confirm that robustness in medical vision–language models emerges from the synergistic interaction of multiple design components rather than from any single mechanism in isolation. These findings reinforce the central premise of this work: robustness should be explicitly modeled during pre-training, rather than treated as an implicit byproduct of reconstruction objectives.

### 4.3.5 Stability Analysis Across Perturbation Levels

Rather than selecting reconstruction layers (as in the reference), we analyze performance stability as a function of perturbation strength. Figure 2 provides a quantitative analysis of robustness by examining model performance as a function of perturbation severity. As perturbation intensity increases, all evaluated methods experience a gradual decrease in VQA-RAD accuracy, reflecting the inherent difficulty of reasoning under degraded visual and textual conditions. However, the proposed method consistently maintains higher accuracy across all perturbation levels and demonstrates a noticeably slower performance decay. This behavior suggests that robustness-aware masked reconstruction encourages the learning of representations that are less sensitive to spurious visual noise and incomplete textual cues. The results further support the hypothesis that explicitly incorporating robustness-oriented objectives during pre-training leads to more stable and transferable medical vision–language representations in the presence of domain shift.

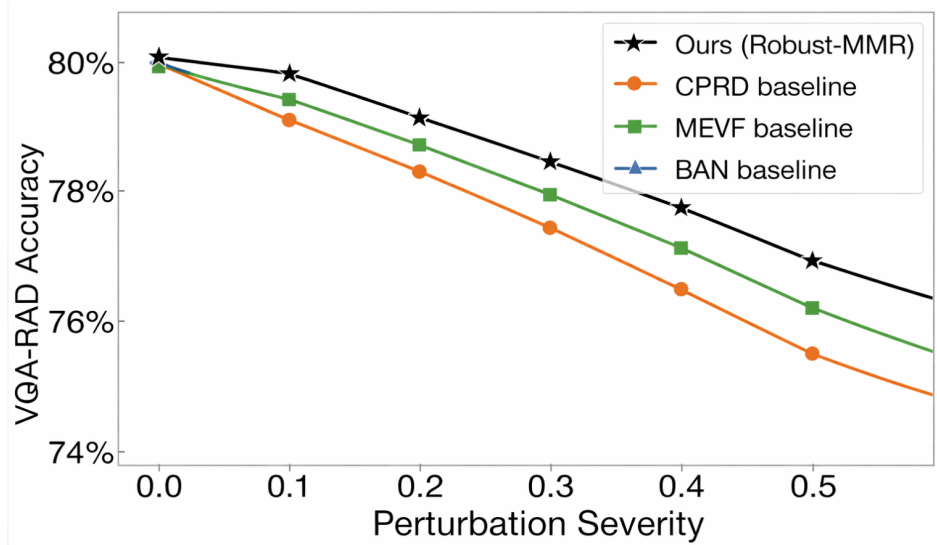

Figure 2. Accuracy stability under increasing perturbation intensity on VQA-RAD. It illustrates the effect of progressively increasing perturbation severity on medical visual question answering performance. Perturbations are applied to both visual inputs (e.g., noise injection and partial masking) and textual inputs (e.g., truncation), simulating realistic degradation encountered under domain shift. The proposed robustness-aware masked reconstruction framework exhibits a slower decline in accuracy compared to representative baseline methods, indicating improved stability under increasingly corrupted inputs.

## 4.4 Qualitative Analysis

To further examine the qualitative behavior of the proposed framework, we present representative Med-VQA examples comparing the baseline MMR model with the proposed Robust-MMR model, as illustrated in Fig. 3. These examples are selected to highlight scenarios in which robustness-aware pre-training enables more reliable cross-modal reasoning under clinically challenging conditions.

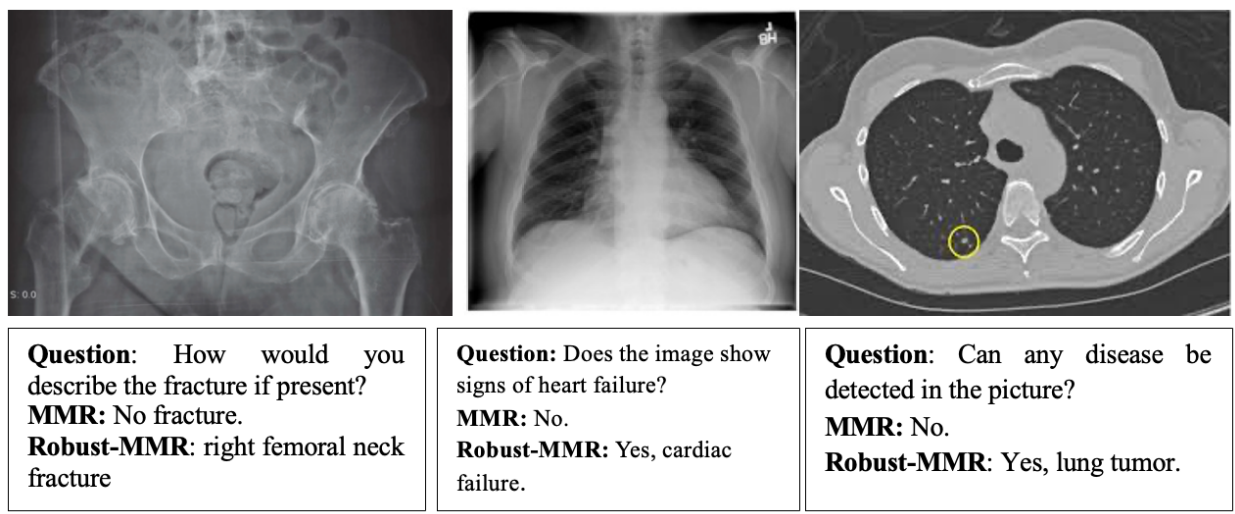

Figure 4. Qualitative Med-VQA examples comparing MMR and Robust-MMR.

In the first example, the question queries the presence and description of a fracture. While the baseline MMR model fails to detect any abnormality, Robust-MMR correctly identifies findings consistent with a **right femoral neck fracture**. This case illustrates the improved sensitivity of Robust-MMR to localized structural abnormalities, likely attributable to its robustness-oriented masked reconstruction strategy. In the second example, the question asks whether the image shows signs of heart failure. The baseline model answers negatively, whereas Robust-MMR identifies imaging findings **suggestive of cardiac failure**. Rather than relying on superficial visual cues, Robust-MMR appears to integrate global anatomical context, demonstrating improved reasoning for conditions that require holistic interpretation of the image. In the third example, the question concerns disease detection. The baseline MMR model again produces a negative response, while Robust-MMR detects a **lung lesion consistent with a tumor**. This example highlights the ability of Robust-MMR to associate subtle visual patterns with disease-related textual concepts, even when such patterns may be partially obscured or degraded.

Overall, these qualitative results suggest that robustness-aware masked reconstruction facilitates the learning of more fine-grained and clinically meaningful vision–language associations. By explicitly encouraging stability under perturbation and domain variation during pre-training, Robust-MMR produces answers that are more consistent with expert-level interpretation, particularly for disease detection and structural abnormality assessment.

## 5. Discussion

This study demonstrates that explicitly modeling robustness during pre-training is critical for advancing medical vision–language models beyond controlled benchmark settings. While prior work has shown that masked reconstruction can effectively align visual and textual representations, our results indicate that reconstruction-centric objectives alone are insufficient to ensure stable performance under domain shift. The proposed Robust-MMR framework addresses this gap by integrating robustness-oriented mechanisms directly into the pre-training process, leading to more transferable and resilient representations.

A key insight from our experiments is that robustness gains manifest most clearly under challenging evaluation conditions, such as cross-domain transfer and perturbed inputs. Across medical VQA, image–text classification, and retrieval tasks, Robust-MMR consistently exhibited smaller performance degradation compared to strong baselines. This pattern suggests that robustness-aware pre-training primarily improves the structure of learned representations rather than merely increasing task-specific accuracy. In practical terms, the model learns to rely less on dataset-specific visual artifacts or stylistic language patterns and instead captures semantically meaningful associations that generalize across institutions and acquisition settings [6]. The ablation analyses further highlight that robustness emerges from the interaction of multiple design components, rather than from any single mechanism in isolation. Robust masking encourages reasoning under partial observability, domain-consistency regularization mitigates sensitivity to acquisition and reporting variations, and modality-resilience constraints reduce dependence on a single modality. Together, these components transform masked reconstruction from a signal-recovery task into a robustness-driven learning objective. This finding reinforces the view that robustness should be treated as a first-class design principle in medical foundation models [46].

From a clinical perspective, the qualitative results are particularly informative. Robust-MMR demonstrates improved reasoning for disease detection and structural abnormality assessment when inputs are noisy, incomplete, or ambiguous—conditions that commonly arise in real-world medical data. While the model is not intended to replace clinical judgment, its improved stability under degraded conditions is a necessary prerequisite for safe and reliable deployment in clinical decision-support systems [43]. Despite these strengths, several limitations warrant discussion. First, robustness is evaluated using synthetic perturbations and cross-dataset splits, which, while informative, may not fully capture the complexity of real-world clinical variation. Future work could extend this evaluation to multi-center datasets with explicitly annotated scanner and institutional metadata. Second, Robust-MMR introduces additional regularization terms and perturbation strategies that modestly increase training complexity. Exploring more efficient formulations or adaptive robustness mechanisms remains an important direction for future research.

Finally, although this work focuses on medical vision–language tasks, the underlying principles of robustness-aware masked reconstruction are broadly applicable. Similar strategies may benefit other multi-modal learning scenarios characterized by heterogeneous data sources and incomplete observations, such as pathology, genomics–imaging integration, and longitudinal electronic health record analysis.

In summary, this work provides both empirical evidence and conceptual motivation for integrating robustness objectives directly into multi-modal pre-training. By shifting the role of masked reconstruction from pure reconstruction fidelity toward stability and invariance, Robust-MMR offers a promising step toward more reliable medical vision–language models suitable for real-world clinical environments.

# 6. Conclusion

In this work, we addressed a fundamental limitation of existing medical vision–language pre-training methods: their vulnerability to domain shift arising from heterogeneous imaging sources, institutional practices, and reporting styles. While masked reconstruction has proven effective for learning joint visual–textual representations, we showed that reconstruction alone is insufficient to guarantee robustness in real-world clinical settings. To this end, we proposed Robust Multi-Modal Masked Reconstruction (Robust-MMR), a robustness-aware pre-training framework that integrates asymmetric perturbation-aware masking, domain-consistency regularization, and modality-resilience constraints into masked reconstruction–based learning. By explicitly modeling robustness during pre-training, Robust-MMR encourages the learning of domain-invariant and modality-resilient representations rather than relying on downstream fine-tuning to compensate for distributional mismatch.

Extensive experiments across multiple medical vision–language tasks demonstrated the effectiveness of the proposed approach. Robust-MMR consistently outperformed state-of-the-art baselines under cross-domain and perturbed evaluation settings while maintaining competitive in-domain performance. The improvements were observed across medical visual question answering, cross-domain image–text classification, and robust image–caption retrieval tasks, highlighting the general applicability of the framework. Qualitative analyses further confirmed that robustness-aware pre-training leads to more reliable reasoning for disease detection and structural abnormality assessment under degraded input conditions. Overall, this work emphasizes that robustness should be treated as a first-class objective in medical vision–language pre-training. By aligning masked reconstruction with robustness-oriented design principles, Robust-MMR provides a practical and effective pathway toward deploying vision–language models that are more reliable, transferable, and suitable for heterogeneous clinical environments.